\documentclass{IEEEojcsys}

\usepackage[colorlinks,urlcolor=blue,linkcolor=blue,citecolor=blue]{hyperref}

\usepackage{color,array}

\usepackage{graphicx}

\usepackage{amsmath,amssymb,amsfonts, bm}

\usepackage{algorithm}
\usepackage[noEnd=true,indLines=true]{algpseudocodex}

\usepackage{xcolor}

\graphicspath{{./}{../}}

\begin{document}

\title{Fading Expert Guidance: \linebreak Bridging Model-Based and Learning-Based Control for \linebreak Abortable Autonomous Overtaking}

\author{Jinxiong Lu\affilmark{1}}
\author{Gokhan Alcan\affilmark{2}}
\author{Ville Kyrki\affilmark{1}}

\affil{Department of Electrical Engineering and Automation, Aalto University, Finland}
\affil{Automation Technology and Mechanical Engineering, Faculty of Engineering and Natural Sciences, Tampere University, Finland}

\corresp{Corresponding Author: Gokhan Alcan (e-mail: \href{mailto:gokhan.alcan@tuni.fi}{gokhan.alcan@tuni.fi})}

\authornote{This work was supported by the Research Council of Finland under Grants 345661, 347199 and 370881.}

\markboth{Fading Expert Guidance for Autonomous Overtaking}{J. Lu {\itshape et al}.}

\begin{abstract}
Overtaking on two-lane roads is a safety-critical decision-making problem for autonomous vehicles, since oncoming traffic may force the ego vehicle to abort and merge back to its original lane. Deep reinforcement learning (DRL) is promising for such continuous-control tasks, but it requires substantial interaction data and may produce unsafe exploratory actions during training. Model-based controllers provide structured behavior, but depend on model fidelity and hand-designed logic. This paper proposes a fading expert-guidance framework that uses a model-based controller to guide DRL training for autonomous overtaking. The expert combines a constrained iterative LQR (CiLQR) planner with PID-based auxiliary controllers, activated when the optimized plan becomes infeasible. The expert action enters the actor objective through a fading term, interpreted as a time-varying soft trust region around the expert policy. This term first biases the policy toward the expert and then vanishes, allowing optimization according to the reinforcement-learning objective. The method is evaluated with PPO, TD3, and SAC. Simulations with bidirectional traffic show improved sample efficiency and final task performance. The safety effects are algorithm-dependent: guidance reduces vehicle collisions for SAC, eliminates boundary collisions for TD3, and leaves the PPO collision profile unchanged. Overall, the results support fading expert guidance as a practical mechanism for transferring model-based driving knowledge to learning-based controllers without restricting the final policy to imitation in an abortable overtaking task with bidirectional traffic.
\end{abstract}


\maketitle

\section{Introduction}

Overtaking maneuvers present a challenging problem for autonomous vehicles due to the need to assess and react to the behaviors of other vehicles, as well as the potential risks associated with executing the maneuver. Two-lane rural-style roads with bidirectional traffic are a particularly challenging instance of this problem because the appearance of an oncoming vehicle on the opposite lane is precisely the kind of unexpected, safety-critical event that a learning-enabled controller must handle by aborting and recovering, rather than by completing the planned maneuver.

Traditionally, control engineering tools and heuristic rules have been used for trajectory planning and decision making of autonomous overtaking \cite{Palatti2021}. However, it is difficult to ensure both the coverage of such rules and their efficiency in complex real-world situations. Moreover, their closed-loop performance is bounded by the fidelity of the underlying model and by the engineering effort spent on enumerating corner cases. With the increase in available computational resources and development of AI methods, there has been an increasing trend to study the problem from the perspective of learning.

Deep Reinforcement Learning (DRL) is a powerful tool in highway autonomous driving that helps make complex decisions. Several papers have studied its application in autonomous driving, particularly in the context of highway overtaking. For example, Tang et al. \cite{Tang2022} applied the Soft Actor-Critic algorithm to solve the highway overtaking and planning problem with continuous action space, and Xiaoqiang et al. \cite{Xiaoqiang2022} proposed to incorporate a graph neural network in DRL for highway autonomous driving. However, DRL suffers from its need of a large amount of interaction data for learning, and offers no built-in mechanism to retain the safety guarantees of model-based control during the early, exploratory phase of training.

Recently, it has been proposed to increase the data efficiency of DRL by integrating guidance from an expert system in the learning \cite{Kim2022, Aksjonov2023}. For example, model predictive control \cite{Kim2022} or a fuzzy-logic based expert system \cite{Wu2022} can be utilized to improve the performance of reinforcement learning in autonomous driving. While integrating expert knowledge, for instance through demonstrations or predefined rules, can accelerate DRL \cite{Kim2022, Aksjonov2023, Wu2022}, a common challenge is that the agent's performance may become capped by the proficiency of the expert. If the guidance is too rigid or maintained indefinitely, it can stifle the agent's exploration and prevent it from discovering policies superior to the expert's. Conversely, insufficient guidance may not adequately improve sample efficiency or ensure safety during initial learning phases. Therefore, a critical question, which this paper addresses, is how to dynamically balance expert guidance with the agent's own learning experience to benefit from expert structure early in training, while crucially enabling the agent to ultimately transcend the expert's capabilities in complex decision-making tasks like autonomous overtaking.

To address this, this paper proposes a novel practical framework for abortable autonomous overtaking, in which guidance from an expert system is incorporated into DRL through a fading guidance function. Conceptually, the fading guidance acts as a \emph{time-varying soft trust region} around the expert policy: at the beginning of training the actor is heavily regularized toward the model-based expert, providing strong inductive bias and reducing risky exploration; over time the regularization weight decays toward zero so that, asymptotically, the stationary points of the actor objective coincide with those of the underlying DRL algorithm. This mechanism is designed to initially steer the agent towards structured and effective behaviors, thereby improving sample efficiency, but then systematically reduces the expert's influence. This calibrated fading allows the DRL agent to progressively rely on its own exploration and learning, with the explicit goal of not only matching but also improving beyond the performance of the expert controller. Our approach thus seeks to combine the initial structure and sample efficiency benefits of expert guidance with the potential of DRL to discover highly optimized, and potentially super-expert, policies.

To summarize, the main contributions of this work are:

\begin{itemize}
    \item We propose an expert-guided deep reinforcement learning framework for autonomous overtaking on two-lane roads with bidirectional traffic. The framework uses a model-based expert only during training, allowing the learned policy to operate independently at deployment time.

    \item We design a model-based guidance system that combines a constrained iterative LQR planner with PID-based auxiliary controllers. The CiLQR controller generates nominal overtaking and lane-following behavior, while the auxiliary controllers handle fallback maneuvers such as following the leading vehicle, aborting an overtake, and merging back safely when the optimized plan becomes infeasible.

    \item We introduce a fading guidance mechanism that regularizes the actor toward the expert policy at the beginning of training and gradually reduces this influence over time. This allows the agent to benefit from expert knowledge during early exploration while retaining the ability to deviate from, and potentially improve upon, the expert policy later in training.

    \item We show that the proposed guidance mechanism is compatible with multiple representative continuous-control DRL algorithms by integrating it with PPO, TD3, and SAC. The resulting guided variants are evaluated against both their unguided baselines and the model-based guidance system.

    \item We provide an empirical evaluation showing that fading guidance improves learning efficiency and final driving performance in the considered overtaking scenario. The results further show that the safety impact of guidance depends on the underlying DRL algorithm and collision type: for example, G-SAC substantially reduces vehicle collisions compared with SAC, while G-TD3 eliminates boundary collisions compared with TD3.
\end{itemize}

The paper is organized as follows. Section~II reviews rule-driven, optimization-driven, learning-based, and hybrid model-/learning-based approaches for autonomous-driving decision making and motion planning. Section~III presents the proposed framework, including the simulation environment, the CiLQR--PID model-based expert, and the fading-guidance formulation used to train the DRL agents. Section~IV reports the training and evaluation results for PPO, TD3, SAC, and their guided variants, followed by a safety analysis based on vehicle and boundary collision rates. Section~V concludes the paper and discusses directions for future work.

\section{Related Work}
Approaches to motion planning and decision making in autonomous driving can be broadly grouped into four categories: rule-driven, optimization-driven, learning-based, and hybrid model-/learning-based methods.

\subsection{Rule-Driven Methods}
Finite State Machines (FSMs) are a classical rule-driven approach for real-time autonomous-vehicle control \cite{Bae2020}. For example, an FSM-based highway-driving controller with feedback control laws and state-transition conditions was proposed in \cite{Zhang2017}. Such methods are attractive because their logic is interpretable and can be directly linked to driving maneuvers. However, rule-driven systems are difficult to scale to complex traffic scenarios, since manually designed transition rules may fail to cover rare or safety-critical corner cases. Moreover, the design and tuning of these rules require substantial engineering effort.

\subsection{Optimization-Driven Methods}
Motion planning can also be formulated as an optimal-control problem. Optimization-driven methods are advantageous because they can incorporate multiple objectives and constraints within a unified formulation \cite{Chen2018}. Gradient-descent and B\'{e}zier-curve-based optimization methods have been applied successfully in static-obstacle scenarios \cite{Zhang1999}, although their performance in highly dynamic environments is more limited. Simplified collision-avoidance constraints have also been studied to improve computational efficiency \cite{Luo2020}. Model Predictive Control (MPC) has been widely used for lane-changing and decision-making problems in highway driving, including settings with uncertainty in future traffic scenarios \cite{Camacho2007, Cesari2017}. The iterative Linear Quadratic Regulator (iLQR) is another optimization-driven control method for dynamic systems \cite{Bemporad2002}. However standard iLQR does not directly handle constraints such as obstacle avoidance and actuator limits. To address this issue, constrained iLQR (CiLQR) was introduced for constrained optimal control of nonlinear systems \cite{Chen2019}. Despite these advantages, optimization-driven methods can be computationally demanding, particularly for nonlinear and non-convex planning problems. Their closed-loop performance also depends strongly on the fidelity of the underlying model.

\subsection{Learning-Based Methods}
Learning-based approaches, especially deep reinforcement learning (DRL), have gained increasing attention for autonomous-driving decision making. For example, \cite{Ye2020} proposed an automated lane-change strategy for highway environments using Proximal Policy Optimization, while \cite{Wolf2018} used compact semantic scene descriptions with reinforcement learning to improve behavior generation across different driving scenes. DRL is attractive because it can optimize policies directly from interaction data and can handle high-dimensional observations and continuous action spaces. However, these advantages come with important limitations. In complex environments, DRL often requires large amounts of training data and may converge slowly or become trapped in undesired local optima due to insufficient exploration. Some studies reduce the complexity of the problem by using discrete action spaces \cite{Chakraborty2023, Mo2019}, but such simplifications can limit the ability to execute smooth and precise maneuvers in continuous driving scenarios.

\subsection{Hybrid Model-Based and Learning-Based Methods}
A growing body of research seeks to combine the structure and safety benefits of model-based control with the adaptability of learning-based policies. Guided policy search~\cite{Levine2013_GPS} uses trajectory optimization to supervise policy learning, and its model-predictive variants supervise the policy directly with an MPC or optimal-control expert~\cite{Zhang2016_MPCGPS,Carius2020_MPCNet}. The weight given to such guidance need not be fixed, it has been tuned automatically during learning~\cite{Zhang2016_MPCGPS} and decayed over training so that the expert influence vanishes asymptotically~\cite{Li2025_FadingGuidance}. Residual reinforcement learning~\cite{Silver2018_ResidualRL,Johannink2019_ResidualRL} instead learns a corrective policy on top of a hand-designed controller. In autonomous-driving and robotics applications, expert demonstrations, MPC-generated trajectories, or rule-based policies have also been used to improve the sample efficiency of reinforcement learning~\cite{Kim2022,Wang2023}. Related work has also explored hierarchical policy-guided planning \cite{Rong2020}, combined learning and rule-based decision making \cite{Xu2022}, and environment-model-based reward-machine construction for safe decision making \cite{Chen2023}. These methods are closely related to our objective of using a model-based controller to guide early-stage learning.

Another related direction is safe reinforcement learning, where model-based components are used to restrict or monitor the learned policy. Examples include shielding methods~\cite{Alshiekh2018_Shielding}, control-barrier-function-based and predictive safety filters~\cite{Cheng2019_CBF,Ames2019_CBFTAC,Wabersich2023_SafetyFilter}, and runtime monitoring approaches for learning-enabled systems~\cite{Garcia2015_SafeRLSurvey,Sun2023_RuntimeMonitoring}. These approaches typically aim to enforce safety during policy execution, whereas our guidance system is used primarily during training and is removed at deployment.

The proposed method differs from these approaches in the way expert knowledge is incorporated. Instead of permanently constraining the learned policy to follow, correct, or pass through the model-based controller, we introduce a fading guidance term in the actor loss. This term initially biases the policy toward the CiLQR--PID expert, improving early exploration and sample efficiency, but gradually vanishes so that the final policy is optimized mainly by the reinforcement-learning objective. In this sense, the expert acts as a temporary training guide rather than a permanent deployment-time safety layer. We acknowledge that constrained trajectory optimization, model based policy guidance, and decaying guidance weights are established techniques, and we do not claim a fundamentally new learning mechanism. Our contribution is a practical integration and empirical study of a common fading guidance formulation with PPO, TD3, and SAC, using a CiLQR planner and PID fallback for abortable autonomous overtaking task in bidirectional traffic.

\begin{figure*}[!t]
\centering
\includegraphics[width=\linewidth]{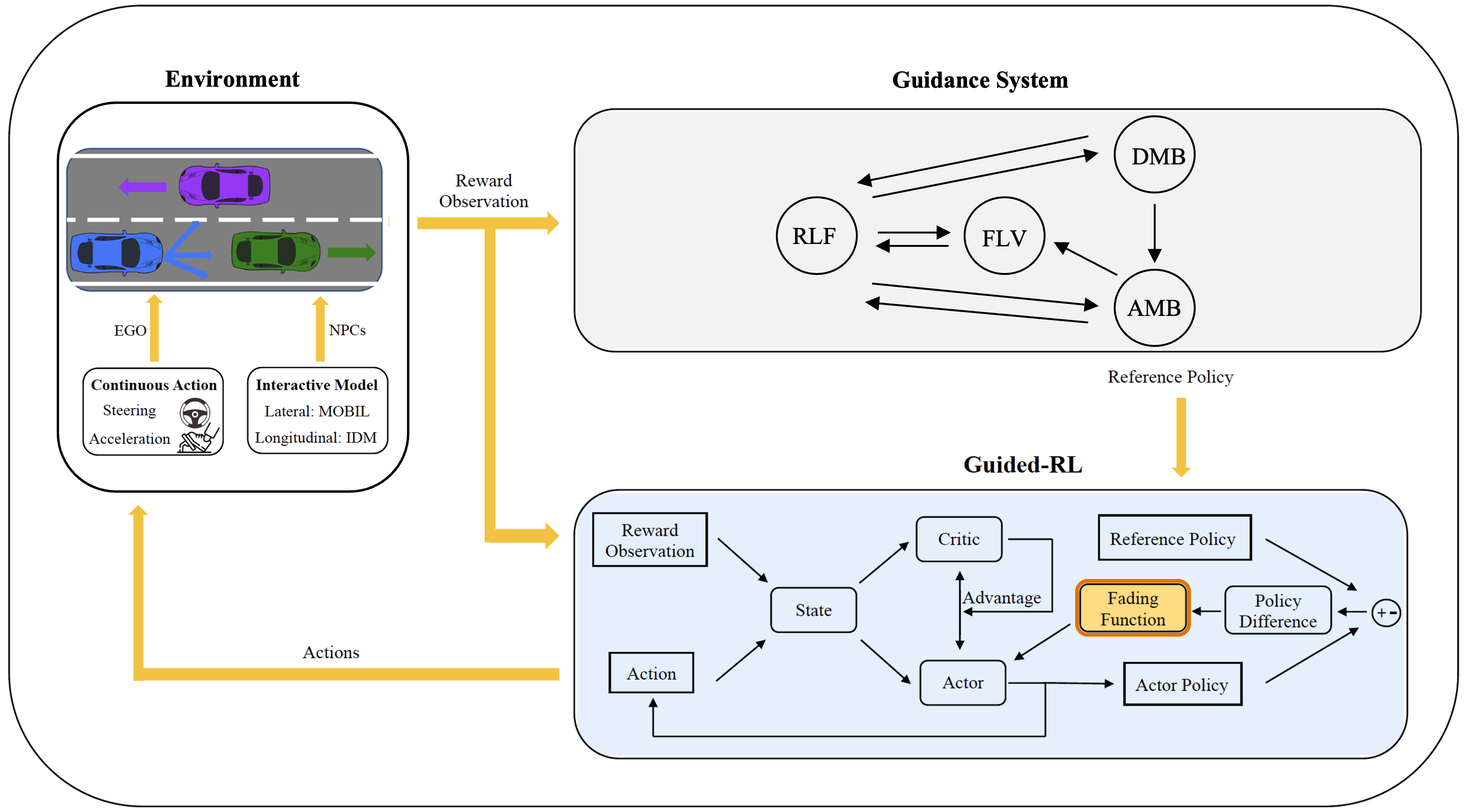}
\caption{An illustration of the integrated DRL and expert system guidance framework. Initial training is heavily influenced by the guidance system, such as the constrained iterative LQR and PID controllers, with the fading guidance function progressively diminishing this expert influence over time. The guidance system (shaded) is used only during training.}
\label{fig1}
\end{figure*}

\section{Proposed Method}

This paper presents a new framework for autonomous driving on two-lane roads with bidirectional traffic that seamlessly blends an expert system into the DRL learning process, using a fading function to progressively diminish the expert system's guidance influence during DRL agent training, as showcased in Fig.~\ref{fig1}. We have implemented our approach on a simulator that takes continuous steering and acceleration inputs to control the ego vehicle mimicking reality. In return, it provides data on the position, velocity, and heading angle of nearby vehicles, as well as immediate rewards. The guidance system is described in detail for reproducibility, but the central mechanism of the framework is the fading guidance term of Section~\ref{subsec:fadingrl}, which sets how strongly and for how long the expert influences learning.

\begin{table*}[!t]
\caption{Parameters of Different Driving Types \label{table1}}
\centering
\resizebox{\linewidth}{!}{%
\begin{tabular}{|*{11}{c|}}
\hline
\shortstack{Driver\\types\\\vphantom{(m/s$^2$)}} &
\shortstack[t]{Desired\\speed\\(m/s)} &
\shortstack[t]{Desired\\time\\(s)} &
\shortstack[t]{Jam\\distance\\(m)} &
\shortstack[t]{Max\\acceleration\\(m/s$^2$)} &
\shortstack[t]{Desired\\deceleration\\(m/s$^2$)} &
\shortstack{Politeness\\\vphantom{acceleration}\\\vphantom{(m/s$^2$)}} &
\shortstack[t]{Safe\\braking\\(m/s$^2$)} &
\shortstack[t]{Acceleration\\threshold\\(m/s$^2$)} &
\shortstack{Length\\\vphantom{acceleration}\\(m)} &
\shortstack{Width\\\vphantom{acceleration}\\(m)} \\
\hline
Timid & 27.8 & 2.0 & 4.0 & 0.8 & -1.0 & 1.0 & 1.0 & 0.2 & 5 & 2 \\
Normal & 33.3 & 1.5 & 2.0 & 1.4 & -2.0 & 0.5 & 2.0 & 0.1 & 5 & 2 \\
Aggressive & 38.9 & 1.0 & 0.0 & 2.0 & -3.0 & 0.0 & 3.0 & 0.0 & 5 & 2 \\
Truck & 23.6 & 2.0 & 4.0 & 0.7 & -2.0 & 1.0 & 1.0 & 0.2 & 6 & 2.5 \\
\hline
\end{tabular}%
}
\end{table*}

A guidance system processes these observations and generates a reference policy, which serves as input for the guided Reinforcement Learning (RL) module. The RL agent employs this reference policy, along with information from the highway environment, to guide the optimization of its Critic and Actor networks, thus promoting efficient learning. Furthermore, a fading function is incorporated to progressively reduce the influence of guidance, enabling the RL agent to learn rapidly in the correct direction initially, and ultimately surpass the guidance it receives. Upon completing the training, the agent no longer relies on the guidance system to execute its policy within the environment.  In essence, the framework functions in three primary phases:
\begin{itemize}
   \item \textbf{Observation Processing:} A guidance system processes the information from the simulator and generates a reference policy. This policy provides the initial direction for the reinforcement learning agent.
   \item \textbf{Guided Learning:} The RL agent uses the reference policy and the highway environment information to optimize its neural networks (e.g. actor/critic). This process is guided by a fading function that gradually reduces the influence of the guidance, thus enabling the RL agent to learn rapidly in the right direction initially and eventually surpass the guidance it receives.
   \item \textbf{Policy Execution:} After the training phase, the RL agent executes its policy within the highway environment, independently of the guidance system.
\end{itemize}

Subsequent subsections provide an in-depth explanation of each component in the framework, including the Environment, Guidance System, and Reinforcement Learning with Fading Guidance.

\subsection{Environment}
\subsubsection{Vehicle Kinematics Model}
The kinematic bicycle model \cite{Polack2017} is used to simulate the motion of the ego vehicle, which is the primary vehicle under the control of the autonomous driving algorithm being studied. This model adeptly captures the critical aspects of the vehicle's movement, striking a balance between simplicity and computational efficiency.

\subsubsection{Non-Player Character Behaviours}
We simulate other vehicles (non-player characters/NPCs) using Intelligent Driver Model (IDM) \cite{Kong2015} for longitudinal motion and Minimizing Overall Braking Induced by Lane change (MOBIL) model \cite{Kesting2007} for lateral motion. Inspired by \cite{Sunberg2018, Kesting2009}, four classes of NPC behavior are simulated (timid, normal, aggressive, truck), each with different IDM/MOBIL parameters, as shown in Table~\ref{table1}. These models react to the ego vehicle only through fixed car-following and lane-changing rules, and neither anticipate its intentions nor adapt to it over time. The results are therefore obtained against reactive but non-anticipatory traffic, and evaluation against interactive or learned driver models is left for future work.

\subsubsection{Observations}
We assume that the ego vehicle is able to observe the positions, headings, and velocities of other nearby vehicles. The observation is represented as an $M \times 6$ matrix whose $i^{th}$ row encapsulates the kinematic data of the $i^{th}$ closest vehicle surrounding the ego vehicle, with the first row ($i=1$) reserved for the ego vehicle itself. This data can be represented as follows:
\begin{equation}
o_i = (p_i, x_i, y_i, v_i^x, v_i^y, h_i)_{i\in\{1,\ldots,M\}}
\end{equation}
In the above equation, $o_i$ signifies the $i^{th}$ row of the observation array. $p_i$ is a boolean variable that indicates the existence of the $i^{th}$ nearest neighbor vehicle. When $p_i$ is zero, all elements within $o_i$ are set to zero instead of removing the entire row from the observation array. This approach is crucial as it ensures a consistent size for the input layer, a requirement for neural networks. The variables $x_i, y_i, v_i^x, v_i^y, h_i$ denote the two-dimensional position, velocity, and heading angle, respectively, of the $i^{th}$ closest neighbor vehicle, as represented in the global coordinate system.

\subsubsection{Actions}
Continuous actions corresponding to acceleration $a$ and steering angle $\delta$ are used to directly set the low-level controls of the vehicle kinematics.

\subsubsection{Reward Function}
We define the ego vehicle's desired behavior using the reward function:

\begin{equation}
    \begin{aligned}
        r=& \; c_1 r_{collision}+c_2  r_{velocity} + c_3 r_{steering}\\
         &+ c_4 r_{acceleration} + c_5 r_{prize}
    \end{aligned}
\label{eqreward}
\end{equation}
where negative rewards $r_{collision}$ are assigned when collisions occur with other vehicles or the ego vehicle crosses highway boundaries. To enhance traffic efficiency and minimize accident risk, a positive reward $r_{velocity}$ is introduced, proportional to the ego vehicle's speed, within the range of maximum and minimum permissible speeds. In an effort to diminish jittering and enhance passenger comfort, a slight penalty $r_{steering}$ is imposed based on the steering angle. A minimal penalty $r_{acceleration}$ is applied in relation to acceleration to promote fuel efficiency. Upon successful arrival at the destination, the agent is granted a substantial reward $r_{prize}$. $\{c_1, ..., c_5\}$ represent the weights assigned to the agent's safety, efficiency, passenger comfort, and task completion.

\subsection{Model-Based Expert with Anomaly-Driven Fallback}
\label{subsec:guidance}

The model-based expert in our framework denotes an expert driver controlled by either a data-driven policy, an optimal control method, or a human demonstrator. For this work, we have designed the expert to act as a high-level decision-maker that operates the ego vehicle within a hierarchical structure. The hierarchy is composed of (i) a primary trajectory-optimization controller that produces a feasible safe plan whenever one exists, and (ii) an \emph{anomaly monitor} that flags the absence of such a feasible plan and switches the closed loop to one of three analytically tractable PID auxiliary controllers. This monitor is the safety-relevant runtime check on the model-based component of the learning-enabled system: it identifies states in which the optimization-based plan is unreliable and routes them to a hand-designed fallback envelope. The decision logic is realized as a Finite State Machine over the (state, transition, action) triples described below.

In order to accommodate the diverse range of maneuvers required by autonomous vehicles, we have incorporated four motion primitives, grouped into primary and auxiliary categories. The primary controller is represented by Responsive Lane Follow (RLF), while the auxiliary controllers consist of Follow Leading Vehicle (FLV), Decelerate and Merge Back (DMB), and Accelerate and Merge Back (AMB). Brief descriptions of these states are provided as follows:

\begin{itemize}
    \item{\textbf{RLF}: In this state, the ego vehicle is instructed to adhere to its assigned trajectory and to commence an overtaking maneuver upon detecting another vehicle. The RLF state is primarily aimed at executing obstacle-free trajectory optimization, and it's expected that the vehicle will transition into this state whenever feasible.}
    \item{\textbf{FLV}: In this state, the ego vehicle is instructed to maintain a specific distance behind the leading vehicle.}
    \item{\textbf{DMB}: During an overtaking maneuver, the ego vehicle is directed to abandon the action if needed. This is done by reducing the speed at a prescribed deceleration rate and rejoining the original lane, provided it is free of any NPC. This is the abort-and-recover mode triggered whenever oncoming traffic invalidates the current overtaking plan.}
    \item{\textbf{AMB}: In the midst of overtaking, the ego vehicle is prompted to expedite the process by accelerating at a given rate and merging back into the lane once it is confirmed to be free of any NPC.}
\end{itemize}

The transitions between those states are triggered as follows:
\begin{itemize}
    \item{\textbf{$\sigma_{RLF}$}: When trajectories planned by RLF do not intersect with either other vehicles or road boundaries. Equivalently, $\sigma_{RLF}$ is true exactly when the constrained trajectory-optimization problem solved by the primary controller is feasible.}
    \item{\textbf{$\sigma_{FLV}$}: When $\sigma_{RLF}$ is False and the ego vehicle is not on the opposite lane.}
    \item{\textbf{$\sigma_{DMB}$}: When $\sigma_{RLF}$ is False, and the ego vehicle is on the opposite lane with its geometric center positioned longitudinally behind the geometric center of the leading vehicle it is attempting to overtake.}
    \item{\textbf{$\sigma_{AMB}$}: When $\sigma_{RLF}$ is False, and the ego vehicle is on the opposite lane with its geometric center positioned longitudinally ahead of the geometric center of the vehicle it is overtaking (i.e., the ego vehicle is in the process of passing or has just passed the other vehicle's center).}
\end{itemize}
The complement $\lnot\sigma_{RLF}$ thus serves as a runtime infeasibility / anomaly signal on the optimization-based plan; when raised, it deterministically routes the closed loop to one of the three PID auxiliary controllers, all of which admit straightforward stability analysis.

\begin{figure*}[!t]
\centering
\includegraphics[width=0.92\linewidth]{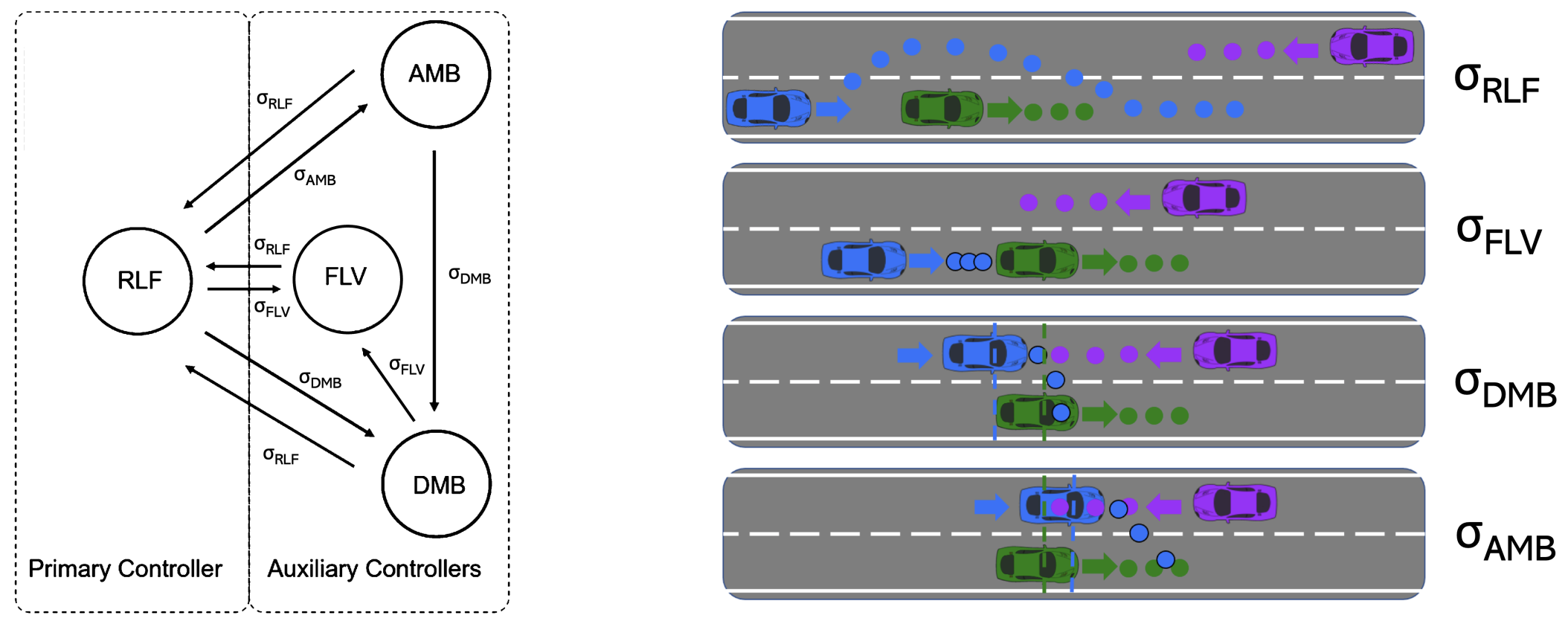}
\caption{FSM schematic of the proposed guidance system (left) and illustration of the four transition conditions $\sigma_{RLF}$, $\sigma_{FLV}$, $\sigma_{DMB}$, and $\sigma_{AMB}$ (right).}
\label{fig2}
\end{figure*}

Fig.~\ref{fig2} provides illustrations of the states and their transition conditions. The implementation of the primary and auxiliary controllers is outlined as follows:

\subsubsection{Primary Controller}

Responsive Lane Follow is considered to be the primary controller and is implemented as a trajectory optimization that solves the following constrained optimal control problem:
\begin{equation}
    \label{eq:opt-problem}
	\begin{aligned}
		\underset{u_0, ..., u_{N-1}}{\text{min}} \quad & \ell^f(s_N)+\sum_{k=0}^{N-1}\ell(s_k, u_k)  \\
		\text{subject to} \quad & s_{k+1}=f(s_k, u_k), \quad k=0,...,N-1\\
        & g(s_k, u_k)<0, \;\;\quad\quad  k=0,...,N-1 \\
        & g(s_N)<0 \\
		\text{given} \quad & s_0=s_{init}
	\end{aligned}
\end{equation}
where $s_k$ and $u_k$ are the state vector and the control input vector at time step $k$, respectively. $N$ is the total number of discrete steps through the horizon of the trajectory. $\ell^f(\cdot)$ and $\ell(\cdot)$ are the final and stage cost functions. $f(\cdot)$ is the bicycle model \cite{Polack2017} based transition dynamics of the ego vehicle. $g(\cdot)$ is the vector of constraints that defines obstacle avoidance and staying within the boundaries of the road.

In order to solve this problem in a computationally efficient way, we employed a Constrained Iterative Linear Quadratic Regulator (CiLQR) algorithm. Here, we summarize only the elements of CiLQR that are needed to make this paper self-contained. The full derivation, regularization scheme, and convergence behavior are documented in \cite{Chen2019,Alcan2022, Alcan2025}. A typical iLQR algorithm solves the problem posed in (\ref{eq:opt-problem}) without the constraints $g(\cdot)$ by employing forward and backward passes iteratively.

To address constraints, we employ the penalty barrier method with exponential barrier function due to its computational efficiency and proved convergence property \cite{Chen2019}. Concretely, each component of the constraint vector is shaped by a differentiable barrier function:
\begin{equation}
c(s, u)=b(g(s, u))
\end{equation}

Since the ideal indicator barrier is non-differentiable, it is approximated by the smooth surrogate
\begin{equation}
b(g(s, u))=q^{b}_{1}\exp\!\big(q^{b}_{2}\,g(s, u)\big)
\end{equation}
where $q^{b}_{1},q^{b}_{2}>0$ are barrier parameters. The exponential penalty was selected because it is smooth and defined on both sides of the constraint boundary, allowing its derivatives to be incorporated directly into the CiLQR backward pass without a separate filtering stage. Its parameters provide a practical tradeoff between constraint penalization and numerical conditioning. Predictive and control-barrier-function-based safety filters are complementary rather than direct substitutes~\cite{Wabersich2023_SafetyFilter}. Their formal guarantees depend on assumptions concerning the model, safe set, invariance, and online feasibility, and their implementation requires an additional optimization layer that may alter the action generated by CiLQR. Integrating such filters with both the guidance controller and the deployed policy would therefore require a different architecture and is left for future work.

Once the constraints are augmented to the cost through the barrier function above, the CiLQR algorithm is implemented through iterations between forward and backward passes until convergence is achieved. For the first iteration, a feasible initial trajectory is designated as the nominal trajectory $\bar{s}, \bar{u}$.

\textbf{Backward Pass} linearizes the system dynamics around the current trajectory, and approximates the cost function ($Q^k(s_k, u_k)$) as quadratic:

\begin{equation}
Q^k(\delta s_k,\delta u_k)
\approx \frac{1}{2}
\begin{bmatrix}
1 \\
\delta s_k \\
\delta u_k
\end{bmatrix} ^\top
\begin{bmatrix}
0 & (Q_s^k)^\top& (Q_u^k)^\top \\
Q_s^k & Q_{ss}^k& Q_{su}^k \\
Q_u^k & Q_{us}^k& Q_{uu}^k
\end{bmatrix}
\begin{bmatrix}
1 \\
\delta s_k \\
\delta u_k
\end{bmatrix}
\end{equation}

\begin{equation}
    \label{eq:Q-derivatives}
    \begin{aligned}
        Q^k_s&=L^k_s+(f^k_s)^\top V^{k+1}_s\\
        Q^k_u&=L^k_u+(f^k_u)^\top V_s^{k+1}\\
        Q^k_{ss}&=L^k_{ss}+(f^k_s)^\top V^{k+1}_{ss}f^k_s+V^{k+1}_sf^k_{ss}\\
        Q^k_{uu}&=L^k_{uu}+(f^k_u)^\top V^{k+1}_{ss}f^k_s+V^{k+1}_sf^k_{uu}\\
        Q^k_{us}&=L^k_{us}+(f^k_u)^\top V^{k+1}_{ss}f^k_s+V^{k+1}_sf^k_{us}
    \end{aligned}
\end{equation}

In the above expression, $L(\cdot)$ represents the cost function augmented with the barrier functions corresponding to constraints. The subscripts in (\ref{eq:Q-derivatives}) indicate either the first or the second derivatives of the function with respect to the variable they annotate. Meanwhile, $V^{k+1}(\cdot)$ denotes the value function for the subsequent iteration ($k+1$). The optimal control at each step is then determined as:
\begin{equation}
\label{eq:ctrl-policy}
\delta u_k^* = -(Q^k_{uu})^{-1}(Q_u^k+Q^k_{us}\delta s_k)
\end{equation}

By plugging this optimal control in the approximated action value function, the derivatives of the value function at time $k$ can be calculated as:
\begin{equation}
    \begin{aligned}
        V_s^k&=Q_s^k-Q_u^k(Q_{uu}^k)^{-1}Q_{us}^k\\
        V^k_{ss}&=Q^k_{ss}-Q^k_{su}(Q^k_{uu})^{-1}Q^k_{us}
    \end{aligned}
\end{equation}

Since $V^{N}(\cdot)$ equals to $\ell^f(\cdot)$, optimal control policy given in (\ref{eq:ctrl-policy}) can be calculated at each step starting from $N$ to 0.

\textbf{Forward Pass} simulates the dynamics forward once the backward pass is completed, by using the optimal control starting from 0 to $N$ as:
\begin{equation}
    \begin{aligned}
        s_0&=s_{init}\\
        u_k&=\bar{u}_k+q_k-(Q^k_{uu})^{-1}Q^k_{us}\delta s_k \\
        s_{k+1}&=f(s_k, u_k)
    \end{aligned}
\end{equation}

where $q_k=-(Q^k_{uu})^{-1}Q^k_u$ is the feedforward term and the remaining term applies linear state feedback around the nominal trajectory, so that $u_k=\bar{u}_k+\delta u_k^*$ with $\delta u_k^*$ from (\ref{eq:ctrl-policy}). Upon the completion of the forward pass, the nominal trajectory $(\bar{s}, \bar{u})$ is replaced by the improved updated trajectory $(s, u)$. These steps repeatedly follow each other, until the convergence is achieved. When the iteration fails to return a feasible trajectory (either because the solver does not converge within its iteration budget, or because the returned trajectory still violates $g(\cdot)$) the FSM raises $\lnot\sigma_{RLF}$ and the closed loop switches to the appropriate PID auxiliary controller, as described next. Divergent or infeasible solves are therefore handled by construction. The reference action supplied to the learning agent is always well defined, and the cost of a failed solve is a more conservative fallback maneuver. This is the runtime anomaly-handling step that keeps the model-based component within a hand-designed fallback envelope. 

\subsubsection{Auxiliary Controllers}

While the primary controller efficiently manages obstacle-free trajectory planning, it lacks the capability to deal with exceptional scenarios, particularly the need to abort an overtaking maneuver. To remedy this, we have introduced three additional motion primitives as auxiliary controllers (FLV, DMB, and AMB). These controllers, detailed at the start of this section, aid in addressing these unique situations, thereby providing a more comprehensive control strategy and were implemented as classical PID controllers as follows:
\begin{equation}
u = K_p  e + K_i  \int e \, dt + K_d  \frac{de}{dt}
\end{equation}
where $K_p$, $K_i$, and $K_d$ are the proportional, integral, and derivative gains, respectively. $e$ is the error defined individually for acceleration and steering inputs in each Auxiliary Controllers.

In FLV, the acceleration error is derived from the spatial discrepancy between the ego vehicle and the leading vehicle. Concurrently, the steering error corresponds to the extent of deviation from the median of the occupied lane. In DMB, the acceleration controller generates a predetermined deceleration rate until the leading NPC is fully surpassed. Subsequent to this event, the error term for the steering controller is determined by the distance from the mid-line of the original lane. Conversely, in AMB, the acceleration controller issues a prescribed acceleration rate until the ego vehicle has completely overtaken the leading NPC. Following this overtaking, the error term for the steering controller is computed based on the distance from the center line of the initial lane.

\subsection{Reinforcement Learning with Fading Guidance}
\label{subsec:fadingrl}
Reinforcement Learning (RL) is a type of machine learning where an agent learns to make decisions by interacting with an environment. It can be formally defined by a tuple $(S, A, P, r, \gamma)$ representing the state space $S$, the action space $A$, the state transition probability $P$, the reward function $r$, and the discount factor $\gamma$. An agent learns from the environment by observing the state  $s \in S$, taking an action $a \in A$, and receiving a reward $r(s, a)$.

An optimal value function $Q^{*}: S \times A \rightarrow \mathbb{R}$ describes the expected cumulative rewards when starting from a given state and following the best policy. The goal of the agent is to learn the optimal policy
\begin{equation}
    \pi^{*}(s)=\underset{a}{\operatorname{argmax}} Q^{*}(s, a)
\label{eq-policy}
\end{equation}
that corresponds to the optimal value function
\begin{equation}
    Q^{*}(s, a)=\max _{\pi} \mathbb{E}\left(\sum_{t=0} \gamma^{t} r\left(s_{t}, a_{t}\right) \mid s_{0}=s, a_{0}=a\right).
\label{eq-valuefcn}
\end{equation}
Throughout the paper, $r$ denotes the instantaneous reward defined in (\ref{eqreward}), cumulative discounted returns appear only inside expectations such as (\ref{eq-valuefcn}).

In our framework, policy from the guidance system guides the RL agent during training, and a fading function is also introduced to fade the guidance effect gradually, which enables the agent to learn in the right direction fast at the beginning and learn beyond the guidance in the end.

The RL agent receives a reference policy from the guidance system, which steers the update of the Actor Networks using the loss function:
\begin{equation}
L_{guidance}=\delta (\pi_{actor}(s), \pi_{reference}(s))
\end{equation}
where $L_{guidance}$ represents the guidance loss, while $\delta$ can be any function representing the divergence between its inputs. In our study, we adopt the Mean Squared Error to define $\delta$.

Fading functions regulate the amount of expert guidance during training. While various decaying functions (e.g., linear, stepwise) could be employed, we selected an exponential decay formulation for $\beta(t)$. This choice offers a smooth transition and allows for flexible shaping of the guidance withdrawal---ranging from rapid to prolonged decay---through its parameters. An exponential decay allows for strong initial guidance which diminishes at a rate that can be tuned to balance initial learning stability with the agent's need to explore independently. In contrast, a linear schedule reduces the guidance weight by the same amount per unit of training time throughout. It therefore withdraws much of the expert bias already in the early phase, where value estimates are inaccurate and expert structure is most useful, while retaining a non-negligible weight until the end of training, where residual bias pulls the converged policy back toward the imperfect expert. The exponential schedule instead concentrates the withdrawal in mid-training and leaves an exponentially small weight at the end, so that the stationary points of the actor objective approach those of the underlying DRL algorithm. Further comparative studies of different decay types could be a subject for future work. In this study, we employed the following inverse exponential function:
\begin{equation}
\beta (t)=\frac{q_1}{\exp(q_2t/T)}
\end{equation}

\begin{figure}[!t]
\centering
\includegraphics[width=0.6\linewidth]{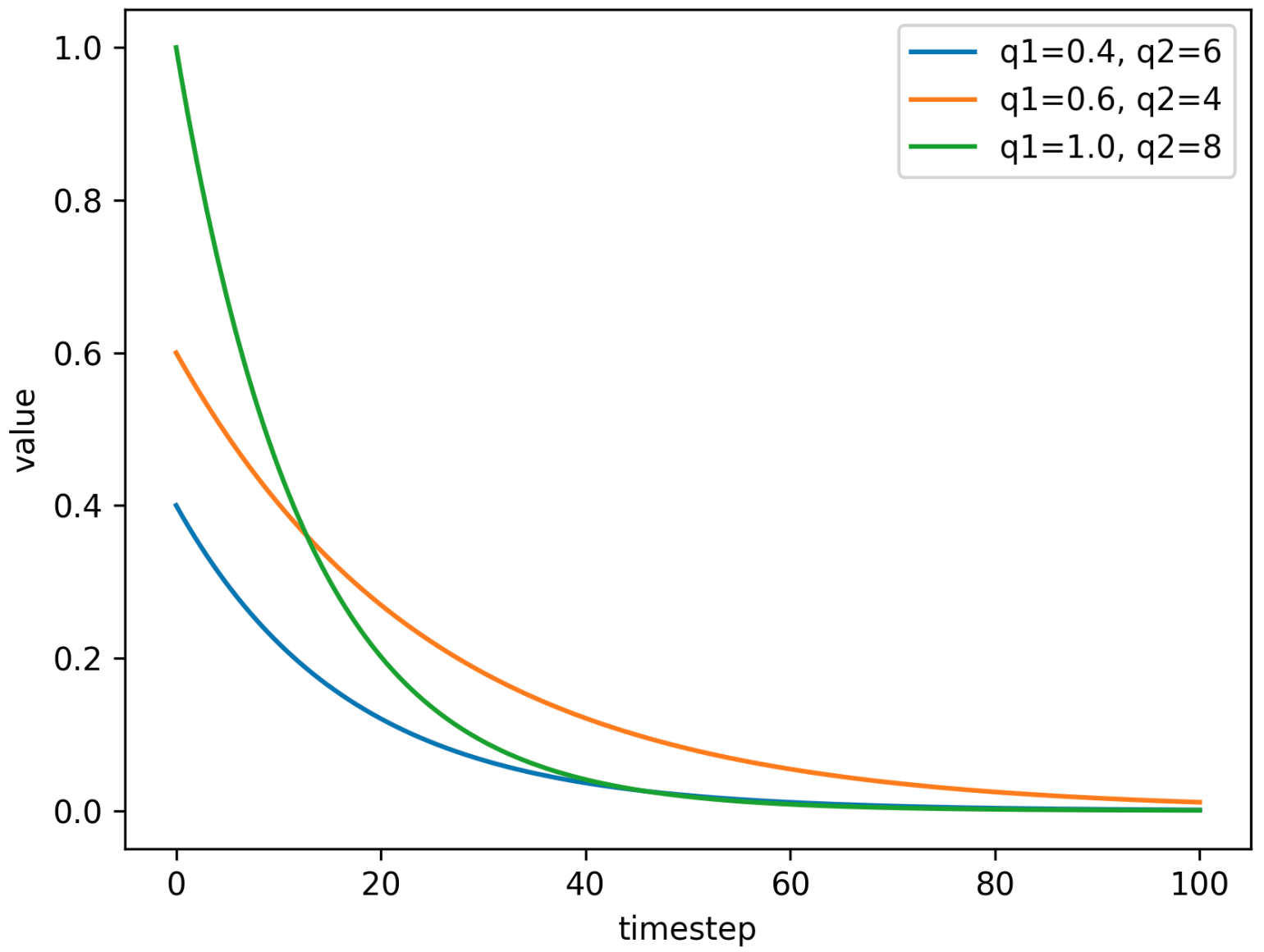}
\caption{Effect of $q_1$, $q_2$ parameters on the shape of fading functions.}
\label{fig3}
\end{figure}

The effect of the parameters $q_1$ and $q_2$ on the shape of the fading function is illustrated in Fig.~\ref{fig3}. The parameter $q_1$ serves as a scaling factor, primarily chosen to ensure that the initial magnitudes of the DRL agent's intrinsic policy loss ($L_{policy}$) and the guidance loss ($L_{guidance}$) are comparable. This balancing is important to prevent either term from dominating the learning process prematurely. The parameter $q_2$ directly modulates the rate at which the expert guidance diminishes over time $t$; its impact on training stability and performance for different DRL algorithms was empirically investigated, as detailed in Section~\ref{sec:TrainingAnalysis} and illustrated in Fig.~\ref{fig4}.

The guidance from the expert system, whose behavior is defined by its underlying rules (FSM, CiLQR, PID controllers), is thus quantitatively integrated into the DRL agent's learning process. The expert's policy, $\pi_{reference}(s)$, provides a target action, and $L_{guidance}$ measures the DRL agent's deviation. The term $\beta(t)L_{guidance}$ in the actor's combined loss function then represents the explicit, weighted influence of the expert's rules on the policy update, acting alongside the agent's own reward-driven $L_{policy}$. As $\beta(t)$ decreases, the direct influence of these expert rules wanes, allowing the agent's learned experience from environmental rewards to predominantly shape its policy.

Lastly, the actor loss function that the optimizer intends to minimize is
\begin{equation}
\label{eq:actor-loss}
L_{actor}=L_{policy}+\beta(t)  L_{guidance}
\end{equation}
where $L_{policy}$ is the policy loss of the specific RL algorithm. When $\beta(t)$ is large, deviations from the expert are strongly penalized and the actor is biased toward the model-based reference policy. As $\beta(t)$ decreases, this penalty weakens, allowing the learned policy to deviate increasingly from the expert when doing so improves the reward-driven objective. Once the training is completed, the guidance system is no longer needed for the agent to execute the trained policy in the environment. The complete procedure, including the CiLQR solve, the anomaly-driven fallback, the fading weight, and the guided actor update, is summarized in Algorithm~\ref{alg:fading}.

\begin{algorithm}[!t]
\caption{Training with Fading Expert Guidance}
\label{alg:fading}
\begin{algorithmic}[1]
\Require training horizon $T$, fading parameters $q_1,q_2$, base DRL algorithm $\mathcal{A}\in\{\text{PPO},\text{TD3},\text{SAC}\}$
\State Initialize actor $\pi_{actor}$, critic(s), and the rollout storage or replay buffer of $\mathcal{A}$
\For{$t=1$ \textbf{to} $T$}
    \State Obtain observation $o_t$ and form the state $s_t$
    \State Solve the constrained problem (\ref{eq:opt-problem}) from $s_t$ with CiLQR
    \If{a feasible trajectory is returned} \Comment{$\sigma_{RLF}$}
        \State $\pi_{reference}(s_t)\gets$ first control of the CiLQR plan
    \Else \Comment{$\lnot\sigma_{RLF}$: anomaly}
        \State Select FLV, DMB, or AMB from the FSM conditions
        \State $\pi_{reference}(s_t)\gets$ output of the selected PID controller
    \EndIf
    \State Execute $a_t$ drawn from $\pi_{actor}(s_t)$, observe $r_t$ and $s_{t+1}$
    \State Store $(s_t,a_t,r_t,s_{t+1},\pi_{reference}(s_t))$
    \If{$t$ is an update step of $\mathcal{A}$}
        \State Sample a batch and update the critic(s) as in $\mathcal{A}$
        \State $L_{policy}\gets$ policy loss of $\mathcal{A}$ on the batch
        \State $L_{guidance}\gets\delta\big(\pi_{actor}(s),\pi_{reference}(s)\big)$ on the batch, using the stored reference actions
        \State $\beta(t)\gets q_1/\exp(q_2 t/T)$
        \State Update $\pi_{actor}$ by minimizing $L_{policy}+\beta(t)L_{guidance}$
    \EndIf
\EndFor
\State \Return $\pi_{actor}$ \Comment{deployed without the guidance system}
\end{algorithmic}
\end{algorithm}

\section{Experimental Results}
\subsection{Simulation}
An open-source autonomous driving simulation framework \cite{Leurent2018} is adapted and customized for the purpose of this research. Our highway environment is characterized by a road length of 1000 m, divided into two lanes. Vehicles in these lanes travel in opposing directions, with each lane having a width of 4 m. The ego vehicle starts with an initial velocity of 45 m/s, has a maximum speed of 60 m/s, and measures 2 m in width and 5 m in length. Its maximum permissible acceleration and steering are 5.5 m/s$^2$ and 1.0 radian, respectively. This speed regime is deliberately aggressive. The closing speed with oncoming traffic is high and the time available to complete or abort an overtake is short, so the abort-and-recover decisions targeted here are triggered frequently and are consequential. The values nevertheless exceed those of typical two-lane rural roads, and the collision rates below characterize this stress-test regime rather than realistic road traffic. The simulation operates at a frequency of 100 Hz and can run for a maximum of 5,000 timesteps per episode. Episodes conclude upon a collision, once the destination is reached, or if the episode's duration exceeds the set threshold. In terms of traffic, the driving types distribution mirrors real-world scenarios with 60$\%$ normal, 20$\%$ timid, 10$\%$ aggressive, and 10$\%$ being trucks \cite{Sunberg2018, Kesting2009}. Forward-driving vehicles are spaced 80 m apart, while vehicles in opposite lanes have 180 m intervals, with random noise added to their base positions. Unlike previous studies, such as \cite{Ye2020}, which only considered vehicle-to-vehicle collisions, our environment also accounts for collisions between the ego vehicle and road boundaries, enhancing realism.

\subsection{Training}
\label{sec:TrainingAnalysis}

We utilized three state-of-the-art algorithms for continuous environments as baselines to validate our method: Proximal Policy Optimization (PPO) \cite{Schulman2017}, Twin Delayed DDPG (TD3) \cite{Dankwa2019}, and Soft Actor-Critic (SAC) \cite{Haarnoja2018}. Within our guided RL framework, these are respectively referred to as G-PPO, G-TD3 and G-SAC. 

The training spanned 800 thousand timesteps with reward parameters set as $c_1$=$-10, c_2$=$1, c_3$=$c_4$=$1$, and $c_5$=$100$. These parameters prioritize reaching the destination as soon as possible, avoiding collisions, and maximizing energy efficiency and stability. From our preliminary experiments, we found the most effective normalization settings to be:
\begin{itemize}
    \item PPO: Both rewards and observations are normalized.
    \item TD3: Only observations are normalized.
    \item SAC: Neither rewards nor observations are normalized.
\end{itemize}

We first analyzed the sensitivity of training performance to the fading schedule. The scaling factor \(q_1\) balances the initial magnitudes of the policy and guidance losses, while \(q_2\) controls how rapidly the guidance coefficient decreases. As shown in Fig.~\ref{fig4}, all three guided algorithms maintain effective learning over the tested range \(q_2\in[4,8]\). Although G-TD3 with \(q_2=8\) and G-SAC with \(q_2=7\) exhibit temporary reward declines, both recover and exceed ten thousand episode reward by the end of training. Smaller values of \(q_2\) produce a gentler transfer from expert guidance to reward-driven learning and yield smoother training curves. This experiment is intentionally a sensitivity analysis of fading rates rather than a comparison with permanent guidance. A constant coefficient would retain a fixed imitation bias throughout training, whereas the tested schedules reduce the coefficient to \(e^{-q_2}\) of its initial value, corresponding to approximately \(1.83\%\) for \(q_2=4\) and \(0.034\%\) for \(q_2=8\). The results therefore demonstrate that the proposed framework remains effective across a broad range of guidance-withdrawal rates and support \(q_2=4\) as a conservative choice that preserves a smoother transition while substantially attenuating the expert regularization by the end of training.

Upon determining optimal fading functions, we delved into the guidance's enhancement on baseline algorithms. For subsequent experiments, $q_1$ was adjusted to match the initial guidance and policy loss magnitudes, and all $q_2$ values were set to 4 for consistent fading and stability. Results were averaged across five random seeds.

Fig.~\ref{fig5} underscores guidance's significant enhancement of the baseline algorithms' sample efficiency:

\begin{figure*}[!t]
\centering
\includegraphics[width=\linewidth]{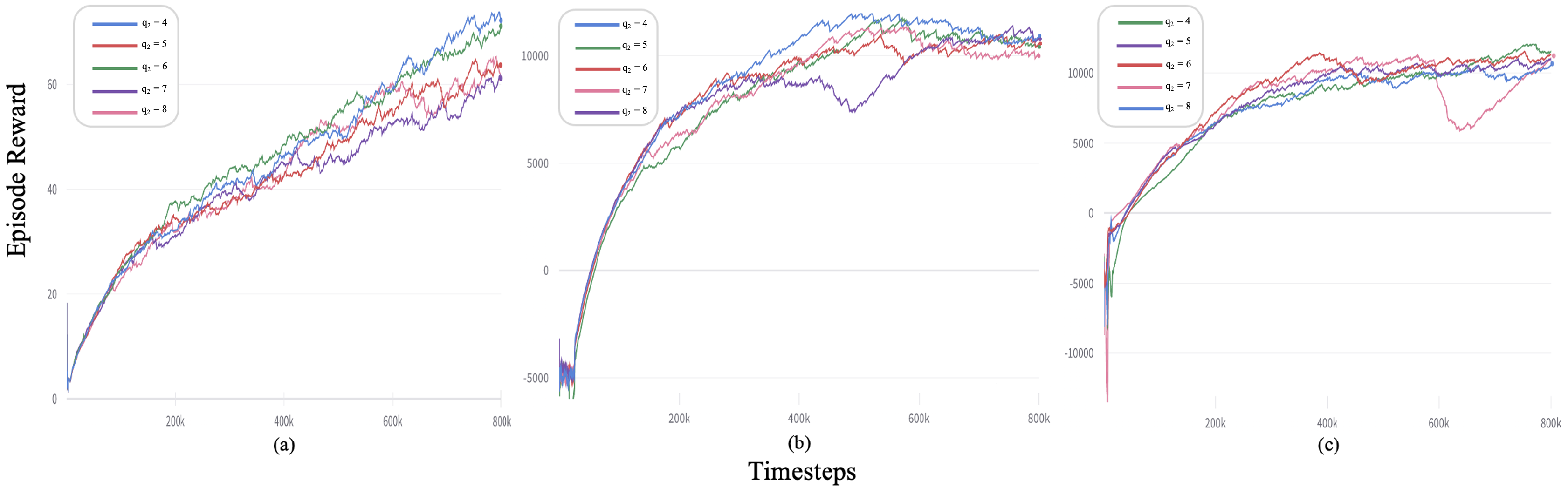}
\caption{Effect of fading functions with different $q_2$ parameters on training performance of the (a) G-PPO, (b) G-TD3, (c) G-SAC algorithms.}
\label{fig4}
\end{figure*}
\begin{figure*}[!t]
\centering
\includegraphics[width=\linewidth]{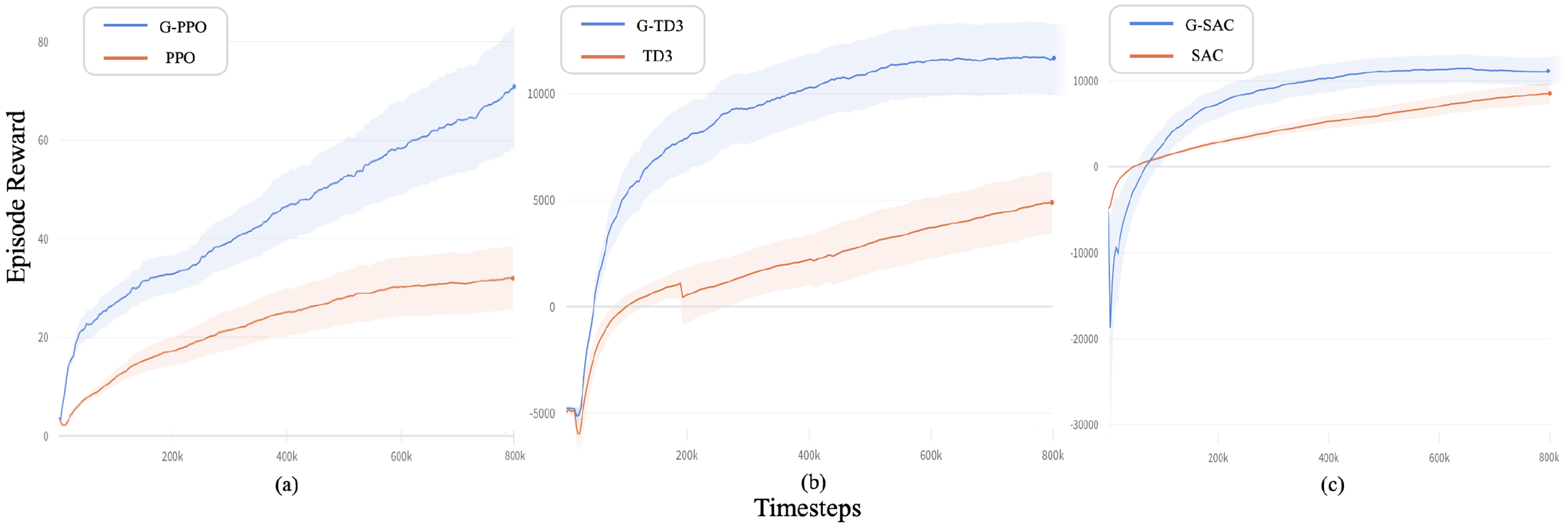}
\caption{Training performance of the three algorithms with and without guidance (average over 5 random seeds).}
\label{fig5}
\end{figure*}

\begin{itemize}
    \item PPO: PPO converges slower than its counterparts. By training's end, while the baseline tends to stabilize, the G-PPO continues its ascent. Here, guidance acts as a potent exploration tool, preventing the RL agent from settling at undesired local minima.
    \item TD3: The G-TD3 converges over 200 thousand timesteps earlier. The baseline reaches only half the episode reward of its guided counterpart. Both exhibit a learning rate shift around 200 thousand timesteps, but the G-TD3 transitions more smoothly.
    \item SAC: Despite initial setbacks, the G-SAC surpasses its baseline before 100 thousand timesteps. Among the three, SAC converges the fastest. Yet, G-SAC further accelerates this, achieving rewards in a quarter of the time taken by its baseline.
\end{itemize}

Fig.~\ref{fig5} evaluates interaction efficiency under a common budget of 800 thousand environment timesteps rather than wall clock efficiency. During training, the guided variants query the expert at every step, whose reported action computation time is approximately \(74\) ms, compared with \(0.10\) to \(0.89\) ms for the learned policies in Table~\ref{tab:table2}. This additional computation is confined to training because the expert is removed after optimization. At deployment, both guided and unguided policies use only their neural networks and produce actions in less than \(1\) ms. The curves therefore show that guidance enables higher rewards within the same interaction budget while preserving the low inference cost of the learned policy, without implying a wall clock training speedup.
\subsection{Evaluation}

In this section, we evaluate the guidance system, baseline RL algorithms, and their guided versions. Given the random noise introduced during training for exploration, all algorithms are anticipated to exhibit enhanced performance during the evaluation phase without action noise.

Beyond the primary metric of rewards, we also dissect the components constituting the overall reward. This granular approach aids in discerning which specific elements the algorithm prioritizes for optimization. Results represent averages over 5 random seeds, with 10 episodes per seed. For a direct comparison during evaluation, rewards remain unnormalized. The findings are detailed in Table~\ref{tab:table2} and Fig.~\ref{fig6}.

\begin{table*}[!t]
\caption{Evaluation Result of Baselines and Guided Algorithms (best results are bolded)\label{tab:table2}}
\centering
\setlength{\tabcolsep}{9.5pt}
\begin{tabular}{|l|c|c|c|c|c|c|c|}
\hline
Evaluation Metrics & Guidance System & PPO & G-PPO & TD3 & G-TD3 & SAC & G-SAC \\
\hline
Episode reward & 5940.9 & 5437.31 & 6421.85 & 3462.42 & 6702.52 & 10112.66 & \textbf{14879.58} \\
\hline
Speed (m/s) & 42.97 & 40.1 & 44.76 & 45.85 & 45.19 & \textbf{59.64} & 57.2 \\
\hline
Displacement (m) & 440.26 & 350.29 & 466.68 & 356.76 & 451.07 & 638.42 & \textbf{807.73} \\
\hline
Computation time (ms) & 74 & \textbf{0.1} & 0.33 & 0.82 & 0.27 & 0.89 & 0.86 \\
\hline
Energy consumption & 1.21 & \textbf{0.34} & 1.39 & 2.45 & 0.98 & 1.36 & 0.52 \\
\hline
Vehicles collision rate & 1 & 1 & 1 & \textbf{0.2} & 0.8 & 0.7 & 0.3 \\
\hline
Boundaries collision rate & \textbf{0} & \textbf{0} & \textbf{0} & 0.8 & \textbf{0} & 0.1 & 0.2 \\
\hline
\end{tabular}
\end{table*}

The evaluation metrics include:

\begin{figure}[!t]
\centering
\includegraphics[width=0.6\linewidth]{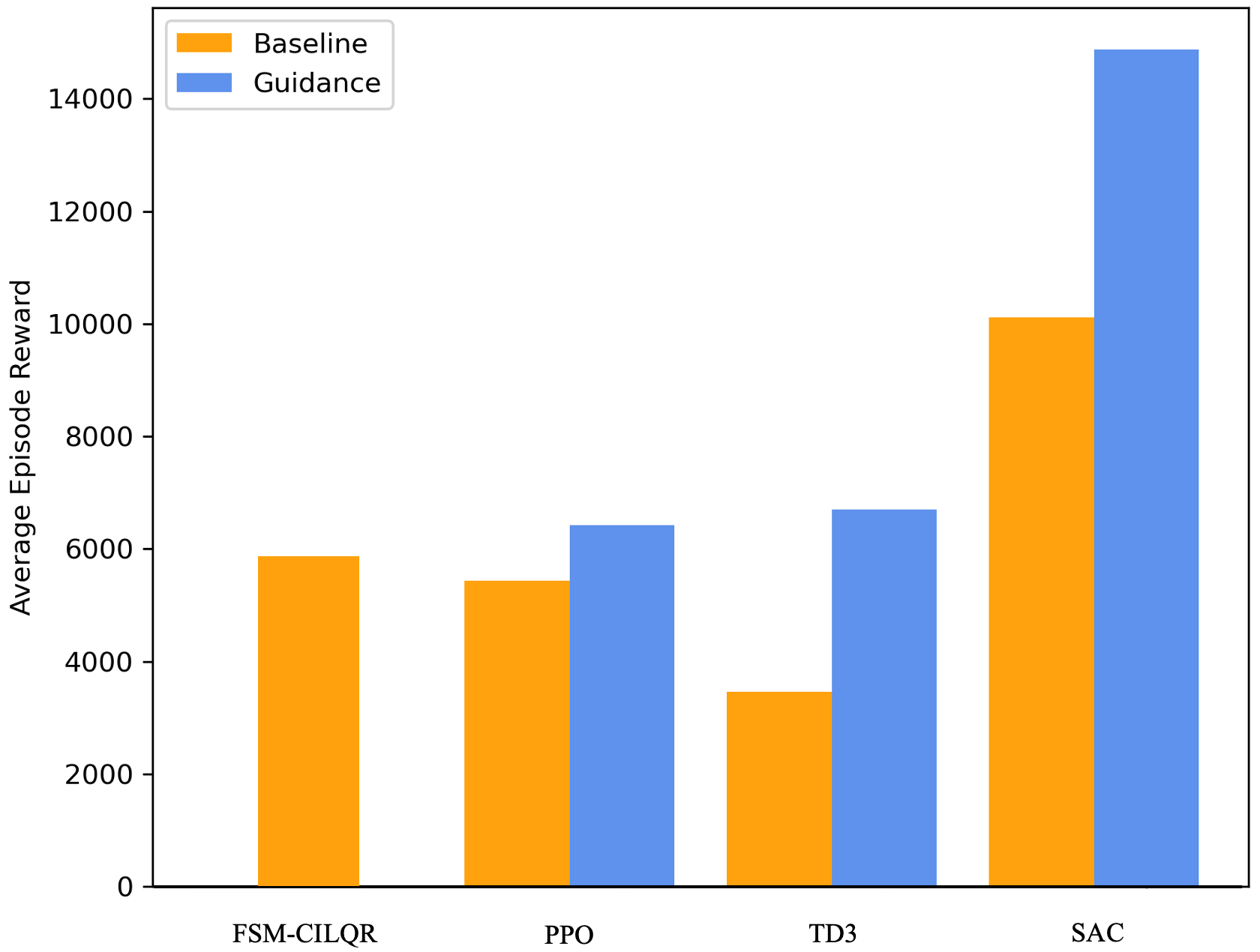}
\caption{Average episode reward of baselines and guided algorithms.}
\label{fig6}
\end{figure}

\begin{itemize}
\item{\textbf{Reward}: Cumulative reward per episode.}
\item{\textbf{Speed (m/s)}: Ego vehicle's average speed per timestep.}
\item{\textbf{Displacement (m)}: Average distance the ego vehicle covers within each episode before termination.}
\item{\textbf{Computation time (ms)}: Average duration required for the agent to determine an action pair from an observation.}
\item{\textbf{Energy consumption}: Represented by the average absolute acceleration per timestep. Longer episodes typically consume more energy, making per-timestep averages more comparable.}
\item{\textbf{Vehicles collision rate}: Average likelihood of the ego vehicle colliding with other vehicles.}
\item{\textbf{Boundaries collision rate}: Average likelihood of the ego vehicle colliding with road boundaries.}
\end{itemize}

Our experimental results affirm the superior performance of guided algorithms over their baselines, particularly in terms of average episode reward---a trend consistent with training outcomes. Guided algorithms consistently achieve higher rewards and displacements, consuming roughly the same computation time. Notably, G-PPO prioritizes speed optimization, while G-TD3 and G-SAC emphasize energy efficiency.

Both G-SAC and G-TD3 agents surpass the guidance system in episode reward, speed, displacement, computation time, and energy consumption; the only metric on which a guided agent does not improve upon the expert is the boundary collision rate of G-SAC. Remarkably, the agents compute actions roughly two orders of magnitude faster, reaching destinations quicker and more energy-efficiently. Instead of manually designing finite state machines to address expert system failures, it's more pragmatic to let the RL agent learn beyond the expert system, aided by a fading guidance function. Experiments validate that as guidance influence diminishes, our algorithm can indeed learn beyond the guidance system.

\subsection{Safety Analysis}
\label{subsec:safety-analysis}

We now discuss the two collision-rate rows of Table~\ref{tab:table2}, namely vehicle collision rate and boundary collision rate, as the primary safety indicators of the closed-loop system. The results show that the effect of fading guidance on these indicators is not uniform across algorithms and collision types.

For PPO, the guided variant preserves the same collision profile as the unguided baseline: both PPO and G-PPO have a vehicle collision rate of $1.0$ and a boundary collision rate of $0.0$. Thus, in this case, guidance improves the episode reward and displacement without degrading the reported collision metrics, but it does not reduce the vehicle collision rate. For TD3, the effect is mixed: G-TD3 reduces the boundary collision rate from $0.8$ to $0.0$, but the vehicle collision rate increases from $0.2$ to $0.8$. For SAC, the opposite trend is observed: G-SAC reduces the vehicle collision rate from $0.7$ to $0.3$, while the boundary collision rate increases slightly from $0.1$ to $0.2$. These results indicate that fading guidance changes the learned policy in a safety-relevant way, but the improvement depends on the underlying DRL algorithm and on which safety failure mode is considered. Each collision rate is estimated from \(50\) evaluation episodes across five seeds, with ten episodes per seed. Under an episode level binomial approximation, the \(95\%\) Wilson intervals have a maximum half width of approximately \(\pm0.14\). The vehicle collision intervals are \([0.11,0.33]\) for TD3 and \([0.67,0.89]\) for \(G_{\mathrm{TD3}}\), while they are \([0.56,0.81]\) for SAC and \([0.19,0.44]\) for \(G_{\mathrm{SAC}}\). The boundary collision intervals are \([0.67,0.89]\) for TD3 and \([0.00,0.07]\) for \(G_{\mathrm{TD3}}\). These separated intervals support clear descriptive differences at the evaluated sample size. By contrast, the boundary collision intervals of \([0.04,0.21]\) for SAC and \([0.11,0.33]\) for \(G_{\mathrm{SAC}}\) overlap, so this difference is not distinguishable at the present resolution. Because episodes produced by the same trained seed may be correlated, the intervals are interpreted as approximate episode level uncertainty estimates rather than independent seed level statistics.

Compared with the model-based guidance system, the learned guided agents substantially reduce vehicle collisions in two out of three cases. The guidance system has a vehicle collision rate of $1.0$, whereas G-TD3 and G-SAC achieve vehicle collision rates of $0.8$ and $0.3$, respectively. In particular, G-SAC provides the strongest overall trade-off among the guided methods: it achieves the highest episode reward, the largest displacement, low energy consumption, and the lowest vehicle collision rate among all guided agents. However, the guidance system, PPO, G-PPO, and G-TD3 all achieve zero boundary collisions, while G-SAC has a nonzero boundary collision rate of $0.2$. The vehicle collision rate of the guidance system should be interpreted in light of its intended role. In the dense and high speed setting considered here, the guidance controller records a vehicle collision rate of \(1.0\) and is therefore not treated as a safety certificate or deployment controller. During training, it provides structured reference actions for lane following, aborting an overtake, and merging back while the learner continues to execute actions and optimize environmental reward. Its value lies in providing an informative early learning bias rather than defining the final policy. The decreasing guidance coefficient progressively relaxes this bias, allowing the learned controller to retain useful structure while departing from the limitations of the expert.

The different effects of guidance across algorithms are consistent with their update mechanisms and preprocessing choices. PPO normalizes both rewards and observations, TD3 normalizes only observations, and SAC normalizes neither. Consequently, the same \(\beta(t)\) does not impose the same relative weighting between \(L_{policy}\) and \(L_{guidance}\) in every algorithm. PPO learns from recent on policy rollouts, whereas TD3 and SAC reuse transitions from replay memory, allowing early guidance to influence a larger number of subsequent updates. SAC also retains an entropy objective that promotes action diversity, while TD3 uses a deterministic actor without a corresponding entropy term. These differences help explain why guidance produces distinct collision profiles across the three algorithms while consistently improving the reported mean reward and displacement.

The main safety conclusion is that fading guidance can improve safety-relevant behavior, especially by reducing vehicle collisions for SAC and eliminating boundary collisions for TD3, while also improving task performance. However, the results also reveal a trade-off between different types of safety violations. This suggests that the guidance loss and reward design influence not only sample efficiency and final reward, but also the distribution of failure modes. Future work will jointly tune the fading schedule and reward weights using collision based objectives and evaluate the resulting policies over larger samples to obtain more precise collision rate estimates.

\section{Conclusion}

This paper presented a fading expert-guidance framework for autonomous overtaking on two-lane roads with bidirectional traffic. The proposed method uses a CiLQR--PID model-based expert to guide DRL during training, while a fading guidance term gradually reduces the influence of the expert in the actor loss. As a result, the learned policy can benefit from model-based guidance during early exploration but operate independently after training.

The framework was evaluated with PPO, TD3, and SAC. The guided variants improved sample efficiency and achieved higher final task performance than their unguided baselines in the considered overtaking scenario. The safety results were more algorithm-dependent: guidance reduced vehicle collisions for SAC and eliminated boundary collisions for TD3, while PPO showed no change in the reported collision profile. These findings indicate that fading guidance can improve safety-relevant behavior, but it does not provide a formal or uniform safety guarantee.

Future work will examine joint tuning of the fading schedule and reward function together with direct comparisons against constant guidance. The framework will also be evaluated with interactive traffic, realistic speed ranges, and broader sample sets, while formal safety mechanisms will be considered as complementary deployment safeguards rather than replacements for fading guidance.

\section*{Acknowledgment}
This work was supported by the Research Council of Finland under Grants 345661, 347199 and 370881. The authors would also like to thank Kevin Sebastian Luck of Vrije Universiteit Amsterdam for his help and support in this work.

\section*{References}

\def\refname{}

\begin{IEEEbiography}
[{\includegraphics[width=1in,height=1.25in,clip,keepaspectratio]{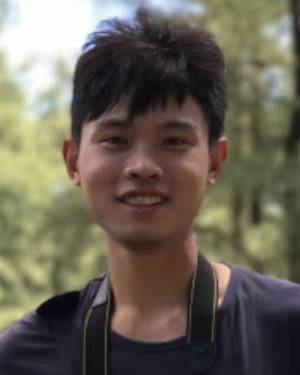}}]{Jinxiong Lu}{\space} received his B.Sc. in Mechanical Engineering from Jilin University, China in 2021. He received his M.Sc. degree with the Department of Electrical Engineering and Automation at Aalto University, Finland. His research focuses on Optimal Control and Reinforcement Learning.
\end{IEEEbiography}

\begin{IEEEbiography}
[{\includegraphics[width=1in,height=1.25in,clip,keepaspectratio]{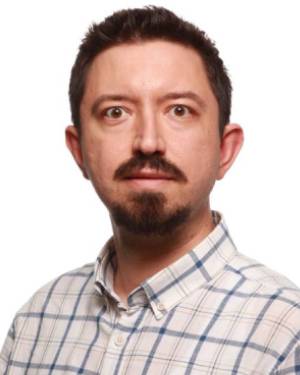}}]{Gokhan Alcan}{\space} received his M.Sc. and Ph.D. \linebreak degrees in Mechatronics Engineering from Sabanci University, Istanbul, Turkey in 2015 and 2019, \linebreak respectively. 

In 2019, he was a postdoctoral researcher in the Control, Vision, and Robotics Research Group at Sabanci University. In 2020-2023, he was a postdoctoral researcher in the Intelligent Robotics Research Group at Aalto University. 

He has been an Assistant Professor of Robotics and Machine Learning at Tampere University, Finland, since 2024, and an Academy Research Fellow funded by the Research Council of Finland since 2025.

His primary research interests are safe model predictive control, constrained optimal control theory, system identification/machine learning, and their applications to dynamical systems.
\end{IEEEbiography}

\begin{IEEEbiography}
[{\includegraphics[width=1in,height=1.25in,clip,keepaspectratio]{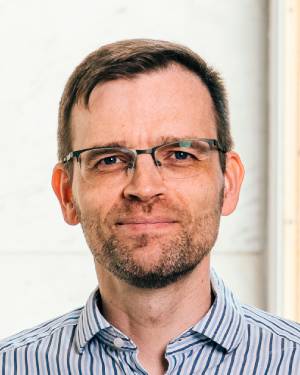}}]{Ville Kyrki}{\space} received the M.Sc. and Ph.D. degrees in computer science from Lappeenranta University of Technology, Lappeenranta, Finland, in 1999 and 2002, respectively.

In 2003-2004, he was a Postdoctoral Fellow with the Royal Institute of Technology, Stockholm, Sweden, after which he returned to Lappeenranta University of Technology, holding various positions in 2003-2009. During 2009-2012, he was a Professor of computer science, Lappeenranta University of Technology. In 2012, he moved to Aalto University, Helsinki, Finland, where he is currently tenured Professor of Intelligent Mobile Machines. His primary research interests include robotic perception, decision making, and learning.

Dr. Kyrki is a Fellow of Academy of Engineering Sciences (Finland), and a member of Finnish Robotics Society and Finnish Society of Automation. He was a Chair and Vice Chair of the IEEE Finland Section Jt. Chapter of CS, RA, and SMC Societies in 2012-2015 and 2015-2016, respectively, Treasurer of IEEE Finland Section in 2012-2013, and Co-Chair of IEEE RAS TC in Computer and Robot Vision in 2009-2013. He was an Associate Editor for the IEEE Transactions on Robotics in 2014-2017.
\end{IEEEbiography}

\end{document}